\documentclass[sigconf]{acmart}

\usepackage{amsfonts}
\usepackage{balance}
\DeclareMathOperator{\argmin}{argmin}
\usepackage{multirow}
\usepackage[ruled,vlined]{algorithm2e}

\AtBeginDocument{%
  }

\copyrightyear{2025}
\acmYear{2025}
\setcopyright{acmlicensed}\acmConference[WSDM '25]{Proceedings of the Eighteenth ACM International Conference on Web Search and Data Mining}{March 10--14, 2025}{Hannover, Germany}
\acmBooktitle{Proceedings of the Eighteenth ACM International Conference on Web Search and Data Mining (WSDM '25), March 10--14, 2025, Hannover, Germany}
\acmDOI{10.1145/3701551.3703492}
\acmISBN{979-8-4007-1329-3/25/03}

\begin{document}

\title{Hyperdimensional Representation Learning for Node Classification and Link Prediction}

\author{Abhishek Dalvi}
\affiliation{%
  \institution{The Pennsylvania State University}
  \city{University Park}
  \state{Pennsylvania}
  \country{USA}}
\email{abd5811@psu.edu}

\author{Vasant Honavar}
\affiliation{%
  \institution{The Pennsylvania State University}
  \city{University Park}
  \state{Pennsylvania}
  \country{USA}}
\email{vuh14@psu.edu}


\begin{abstract}
  We introduce Hyperdimensional Graph Learner (HDGL), a novel method for node classification and link prediction in graphs.  HDGL maps node  features into a very high-dimensional space (\textit{hyperdimensional} or HD space for short) using the \emph{injectivity} property of node representations in a family of Graph Neural Networks (GNNs)  and then uses HD operators such as \textit{bundling} and \textit{binding} to aggregate information from the local neighborhood of each node yielding latent node representations that can support both node classification and link prediction tasks. HDGL, unlike GNNs that rely on computationally expensive iterative optimization and hyperparameter tuning, requires only a single pass through the data set.  We report results of experiments using widely used benchmark datasets which demonstrate that, on the node classification task, HDGL achieves accuracy that is competitive with that of the state-of-the-art GNN methods  at substantially reduced computational cost; and on the link prediction task, HDGL matches the performance of DeepWalk and related methods, although it falls short of computationally demanding state-of-the-art GNNs.
\end{abstract}

\ccsdesc[500]{Computing methodologies~Learning latent representations}
\ccsdesc[300]{Computing methodologies~Semi-supervised learning settings}

\keywords{Hyperdimensional Computing, Graph Neural Networks, Transductive Learning, Representation Learning}


\maketitle

\section{Introduction}
Social networks, citation networks, molecular, e.g., protein-protein interaction networks, etc. are naturally represented as graphs where nodes represent the entities that make up the network and edges between pairs of nodes encode relationships between entities. For example, in a social network, nodes represent individuals and links denote social connections; in a citation network, nodes represent articles, and (directed) links denote citations; In a protein-protein interaction network, the nodes represent proteins and links represent their pairwise interactions.  Two of the key problems presented by such data include  node classification and link prediction \cite{li2020network,kumar2020link}. The state-of-the-art methods for solving such problems rely on graph representation learning methods \cite{zhou2022network,cai2018comprehensive,xu2021understanding,wang2022survey}. 

The state-of-the-art performance achieved by such methods comes at the expense of high computational costs -- and the large carbon footprints and the attendant adverse environmental impacts \cite{DESISLAVOV2023100857,strubell2020energy,de2023growing,DL_Diminishing_Returns} -- of data-hungry deep learning methods that rely on iterative parameter optimization and hyperparameter tuning.
Against this background, we explore a computationally efficient, one-pass learning algorithm in which the model needs to see each data sample only once, as an alternative to state-of-the-art graph neural networks for node classification and link prediction.

\noindent\textbf{Key Contributions:} We introduce a Hyperdimensional Graph Learner (HDGL), a novel method for node classification and link prediction in graphs. HDGL maps node  features into a very high-dimensional space (\textit{hyperdimensional} or HD space for short) using the \emph{injectivity} property of node representations in a family of Graph Neural Networks (GNNs)  and then uses HD operators such as \textit{bundling} and \textit{binding} to aggregate information from the local neighborhood of each node yielding latent node representations that can support both node classification and link prediction tasks. HDGL provides a one-pass learning alternative to GNNs trained using computationally expensive iterative optimization methods. We report results of extensive experiments using widely used benchmark datasets which demonstrate that, on the node classification task, HDGL achieves accuracy that is competitive with that of the state-of-the-art GNN methods at substantially reduced computational cost; and on the link prediction task, HDGL matches the performance of DeepWalk and related methods, although it falls short of computationally demanding state-of-the-art GNN methods. We further show that HDGL, which does not require expensive iterative learning procedures, is well-suited for data/class-incremental learning, making it an attractive alternative to GNNs.

\section{Related Work}

\noindent\textbf{Graph Neural Networks} (GNNs), introduced by \citet{Starita_graphs,Frasconi1998,Gori2005ANM,Scarselli2009}, which learn embeddings of nodes in a graph by aggregating information from local neighborhoods of the nodes, have recently emerged as the dominant approach to graph representation learning \cite{kipf2016semi,veličković2018graph,hamilton2017inductive},  because of their superior performance across  a broad spectrum of applications \cite{wu2020comprehensive,zhang2018network,zhou2022network}. However, as noted above, their state-of-the-art performance comes at a high computational cost and adverse environmental impact. In contrast, HDGL needs only a single pass through the training data, and hence, offers a computationally efficient, low energy footprint alternative to GNN for node classification and link prediction.

\noindent\textbf{Hyperdimensional computing} \cite{KanervaSDM,HD_computing_intro,neubert2019introduction,thomas2021theoretical} offers a brain-inspired alternative to deep neural networks for artificial intelligence and machine learning applications, that maps each object using a sufficiently high-dimensional random encoding (HD encoding for short) to produce a binary or bipolar vector \cite{KanervaSDM,HD_computing_intro}. Simple operations, like element-wise additions and dot products \cite{KanervaSDM,neubert2019introduction,thomas2021theoretical}, are used to perform computations on the encoded objects. Computations on HD vectors, because they yield binary or bipolar vectors, can be realized using low-precision, fast, low-power, energy-efficient hardware \cite{DBLP:journals/csur/KleykoROR23,DBLP:journals/csur/KleykoROR23a}. Hence, there is a growing interest in the use of HD computing in machine learning \cite{neubert2019introduction,DBLP:journals/csur/KleykoROR23a,ge2020classification,morris2022HyDREA}. 

\noindent\textbf{Hyperdimensional Graph Encodings}.  \citet{Poduval_GraphHD} and \citet{Nunes2022GraphHDEG} have investigated 
hyperdimensional representations of graphs for similarity-based retrieval and graph classification. \citet{Nunes2022GraphHDEG} constructs the Graph HD vector using node features derived using PageRank \cite{sergey1998_Pagerank}. \citet{Poduval_GraphHD} assign random HD-vectors to nodes and considers them as node features to build a Graph HD vector. Thus, both methods ignore node features, whereas our focus is on methods that learn node-level representations for node classification and link prediction.

\noindent\textbf{Hyperdimensional Node Embeddings}. Recent work has explored hyperdimensional embeddings of nodes in a graph \citet{Tajana_RelHD,Li_HyperNode}. Specifically, \citet{Tajana_RelHD} proposed RelHD, a novel processing-in-memory (PIM) hardware architecture \cite{mutlu2022modern_PIM} based on FeFET technology \cite{Beyer2020FeFETAV}. \citet{Li_HyperNode} introduced the HyperNode method for learning HD embeddings of nodes. However, both RelHD and HyperNode encode node features as binary indicator variables before mapping them to HD space and hence are unsuited for graphs with continuous or integer-valued node attributes e.g., word counts.
Furthermore, none of the aforementioned methods \cite{Nunes2022GraphHDEG, Poduval_GraphHD,Tajana_RelHD, Li_HyperNode} support link prediction on graphs.

In contrast, HDGL offers an effective and efficient method for constructing more expressive node embeddings that can accommodate not only binary or multi-valued, but also integer-valued or continuous node features. The resulting HD encodings of nodes support both node classification and link prediction.

\section{Preliminaries}
\label{sec:bg}

We proceed to briefly summarize  the key concepts needed to set the stage for introducing HDGL.
\subsection{Node Classification and Link Prediction on Graphs}
Let $\mathcal{G} = (\mathcal{V}, \mathcal{E})$, where $\mathcal{V}$ represents set of nodes, $\mathcal{E}$ represents the set of undirected edges $\mathcal{E} \subseteq \{ (u,v) | u, v, \in \mathcal{V} \}$. Let $N = |\mathcal{V}|$ denote the number of nodes in the graph. Supposed each node $v \in \mathcal{V}$ is described by a tuple of $d$ features $\mathbf{x}_{v} \in \mathbb{R}^{d}$. The $k$-hop neighbors of node $v$ in graph $\mathcal{G}$ are represented using multi-set $\mathcal{N}^{k}(v)$.
The node classification problem can be described as follows: Given labels $y_{w} \in \{1, \cdots, L\}$ for each node $w \in \mathcal{W}$, where $\mathcal{W} \subset \mathcal{V}$; the task is to predict the labels of unlabeled nodes, i.e., nodes in $\mathcal{V} \setminus \mathcal{W}$. Similarly, the link prediction problem entails predicting the missing edges/new edges in $\mathcal{G}$, e.g., predicting new links/recommendation in a social network. We aim to solve both problems by learning a latent representation $\phi(v)$ for each node $v$, utilizing the structure of $\mathcal{G}$, i.e., the connectivity between nodes, the feature vector $\mathbf{x}_v$, and the node labels $y_w$.

\subsection{Hyperdimensional Computing}\label{HD-comp_background}

Hyper-dimensional computing, originally introduced by \citet{KanervaSDM} is a brain-inspired approach to representing information that encodes each object $x \in \mathcal{X}$ using a high dimensional mapping $\phi: \mathcal{X} \rightarrow \mathcal{H}$,  where $\mathcal{H}$ is typically $\{-1,1\}^n$ or $\{0,1\}^n$ where $n$ is sufficiently large \cite{KanervaSDM,HD_computing_intro}. 
All computations are performed in $\mathcal{H}$, using simple operations, e.g., element-wise additions and dot products \cite{KanervaSDM,neubert2019introduction}. The mapping $\phi$ is often random, and low precision which lends itself to fast, low-power, energy-efficient hardware realizations \cite{DBLP:journals/csur/KleykoROR23,DBLP:journals/csur/KleykoROR23a}. In this paper, we assume $\mathcal{H} = \{0,1\}^n$. We proceed to summarize some of the key properties of HD representation.

\noindent
\textbf{Near Orthogonality of Random vectors in HD Space.} HD encodings of objects are produced by  sampling each dimension independently and uniformly from a distribution. Let $\mathbf{a} \in \{0,1\}^{n}$ and $\mathbf{b} \in \{0,1\}^{n}$ be two such HD vectors where $a_{i} \sim Ber(p=0.5); \forall i \in \{1, \cdots ,n\}$ and $b_{i} \sim Ber(p=0.5); \forall i \in \{1, \cdots ,n\} $. We can then say that $\mathbf{a}$ and $\mathbf{b}$ are near orthogonal or   $d_{H}[\mathbf{a},\mathbf{b}]$, the normalized hamming distance between $\mathbf{a}$ and $\mathbf{b}$, $\approx 0.5$ \cite{KanervaSDM}. This is also known as a blessing of HD spaces. 

\noindent
\textbf{High Memory Capacity of HD Representation.}  As noted by \citet{neubert2019introduction}, in the case of an $n$-dimensional binary HD space  where each of the $n$-dimensional vectors is $f$-sparse,  e.g., the number  of nonzero entries is $fn$, the resulting HD space has capacity given by $\binom{n}{\lfloor f \cdot n \rfloor}$. For example, when $n=1000$, and $f=0.05$, the resulting HD space can store approximately $10^{80}$ or almost as many patterns as the  number of atoms in the observable universe (which is estimated to be between $10^{78}$ and $10^{82}$).

\noindent
\textbf{Noise or Error Tolerance.} Due to the distributed representation of data across numerous dimensions in HD spaces, errors in HD computing are limited to only a fraction of bits \cite{ahmad2015properties,neubert2019introduction,thomas2021theoretical}. Thus, HD computing offers  noise tolerance without the need for expensive error correction mechanisms \cite{neubert2019introduction}. 

We next briefly describe some of the key operations used to compute with HD representations:-

\noindent
\textbf{Bundling}, denoted by $\oplus$,  is a bitwise majority operation represented by $\oplus : \{0,1\}^{n} \times \{0,1\}^{n} \rightarrow \{0,1\}^{n}$. Bundling randomly sampled HD vectors results in a new vector which is similar to the input vectors.  Bundling $\oplus$ is associative and commutative i.e $\mathbf{a} \oplus (\mathbf{b} \oplus \mathbf{c}) =  (\mathbf{a} \oplus \mathbf{b}) \oplus \mathbf{c} = (\mathbf{c} \oplus \mathbf{a}) \oplus \mathbf{b}$. Note that in case of Bundling an even number of HD vectors, there is  a possibility of ties since Bundling is a bitwise majority operation. To break ties, a tie-breaking policy needs to be used, e.g., random choice of 0 or 1  on dimensions where ties occur. Bundling can be thought of as the HD space analog of the mean operation euclidean spaces. Bundling, because it commutes, can be used to represent sets as well as multisets of objects encoded using HD vectors.

\vspace*{5pt}
\noindent 
\textbf{Binding}, denoted by $\otimes$, is a bitwise Exclusive OR operation represented by $\otimes : \{0,1\}^{n} \times \{0,1\}^{n} \rightarrow \{0,1\}^{n}$, is an operation which takes two input vectors and maps them to a new vector. An important property of Binding is that Binding two HD vectors results in a vector that is dissimilar to both. Formally, let $\mathbf{a} \in \{0,1\}^{n}$ and $\mathbf{b} \in \{0,1\}^{n}$ be two  sampled HD vectors and $\mathbf{c} = \mathbf{a} \otimes \mathbf{b}$. Due to the near-orthogonality of HD vectors, we can then say that $d_{H}[\mathbf{c},\mathbf{a}] \approx 0.5$ and $d_{H}[\mathbf{c},\mathbf{b}] \approx 0.5$. Binding is invertible, associative and commutative i.e $ \mathbf{b} \otimes (\mathbf{a} \otimes \mathbf{b}) =  (\mathbf{b} \otimes \mathbf{a}) \otimes \mathbf{b} = \mathbf{a} \otimes \mathbf{b} \otimes \mathbf{b} = \mathbf{a}$. Moreover, Binding is distributive over bundling i.e  $\mathbf{a} \otimes (\mathbf{b} \oplus \mathbf{c}) = (\mathbf{a} \otimes \mathbf{b}) \oplus (\mathbf{a} \otimes \mathbf{c})$. Note that Binding also has a reflective property i.e the distance 
between two vectors doesn't change when both vectors are bound to the same vector. Formally,  $d_{H}[\mathbf{a},\mathbf{b}] = d_{H}[\mathbf{c} \otimes \mathbf{a},\mathbf{c} \otimes \mathbf{b}]$. Note that this also works even vectors are slightly different i.e $d_{H}[\mathbf{a},\mathbf{b}] \approx d_{H}[\mathbf{c} \otimes \mathbf{a}',\mathbf{c} \otimes \mathbf{b}]$ if $\mathbf{a} \approx \mathbf{a}'$.

\vspace*{5pt}
\noindent 
\textbf{Rotation}, also called permutation, denoted by $\Pi$, is a bitwise left shift operation $\Pi : \{0,1\}^{n} \rightarrow \{0,1\}^{n}$. A key property of Rotation is that rotation results in dissimilar/orthogonal vectors i.e  $d_{H}[\mathbf{a},\Pi(\mathbf{a})] \approx 0.5$ and it is distributive over both Bundling and Binding and like binding, rotation is also reflective i.e  $d_{H}[\mathbf{a},\mathbf{b}] \approx d_{H}[\Pi(\mathbf{a}), \Pi(\mathbf{b})]$. Rotation was introduced by \citet{gaylerVSA} in order to protect or preserve information because information can be lost/cannot be distinguished because Binding is associative and commutative i.e:- $(\mathbf{a} \otimes \mathbf{b}) \otimes (\mathbf{c} \otimes \mathbf{d}) = (\mathbf{a} \otimes \mathbf{c}) \otimes ( \mathbf{b} \otimes \mathbf{d}) $. In order to make the make a distinction between $(\mathbf{a} \otimes \mathbf{b}) \otimes (\mathbf{c} \otimes \mathbf{d})$ and $(\mathbf{a} \otimes \mathbf{c}) \otimes ( \mathbf{b} \otimes \mathbf{d})$, rotation is used :- $(\mathbf{a} \otimes \mathbf{b}) \otimes \Pi(\mathbf{c} \otimes \mathbf{d})$.

 Note that the operations referenced earlier have equivalents in the bipolar space $\{-1, 1\}^{n}$. Specifically, the Bundle operation can be  represented as signed addition in the bipolar space, while the binding operation aligns with multiplication in the bipolar space.

\subsection{Graph Neural Networks}
GNNs use the node features, together with the connectivity between the nodes information to learn the embedding  $\mathbf{h}_{v}$ of each node $v \in \mathcal{V}$. Most GNN use a neighborhood aggregation scheme where each node recursively aggregates the features of its neighbors to compute its feature vector \cite{GIN_cite,wu2020comprehensive,zhang2018network,zhou2022network}. After $k$ iterations of such aggregation, the node's feature vector yields its embedding. More precisely, the latent representation of node $v$ at the $k$th iteration is given by: $
\mathbf{h}^{(k)}_{v} = \text{Agg}( \{ \mathbf{W}^{(k)}\mathbf{h}^{(k -1)}_{j}; j \in \mathcal{N}^{1}(v) \cup \{ v \} \})
$; where $\{ \cdot \}$ denotes a multiset and $\mathcal{N}^{1}(v)$ denotes the 1-hop neighbors of $v$. Different GNN models, e.g.,  spectral methods \cite{kipf2016semi}, attention-based methods  \cite{veličković2018graph}, and LSTM-based methods \cite{hamilton2017inductive} differ primarily with respect to the  aggregation function they use.

\noindent\textbf{Representational Power of GNN}. The representational power of GNNs can be characterized in terms of their  ability to distinguish between different graph structures \cite{GIN_cite}. The greater the representational power, the more minute the differences between graphs that can be detected by the GNN. GNN have been shown to be at most as powerful as the Weisfeiler-Lehman (WL) graph isomorphism test, which iteratively updates node features by aggregating neighbors' features using an injective aggregation function \cite{weisfeiler1968reduction,huang2021short}. The injectivity of the aggregation function  ensures that the differences in  node neighborhoods result in differences in the resulting aggregations \cite{huang2021short}. Consequently, GNN are at most as powerful as the WL test in distinguishing graph structures if the mapping $\mathcal{A} : \mathcal{G} \rightarrow \mathbb{R}^d$ produced by the iterative aggregation scheme:
$
    \mathbf{h}^{(k)}_{v} = g(\mathbf{h}^{(k -1)}_{v}, f(\{\mathbf{h}^{(k -1)}_{j}; j \in \mathcal{N}^{1}(v) \}))
$ where $g(\cdot)$ and $f(\cdot)$ and $\mathcal{A}$'s graph-level readout are injective.

\section{Hyper-Dimensional Graph Learner}
\label{sec:hdgl}
HDGL constructs a HD representation of graphs that matches the  expressive power of GNN. HDGL exploits the properties of HD operations, namely, Bundling ($\oplus$), Binding ($\otimes$), and Rotation ($\Pi$) to ensure that graphs with similar node features and topologies map to similar HD representations. Specifically, HDGL: 
(i). Maps node features $\mathbf{x}_{v} \in \mathbb{R}^{d}$ to a intermediate HD representation $\mathbf{r}_{v} \in \{0,1\}^{\beta}$ $\forall v \in \mathcal{V}$. 
(ii). Constructs a latent HD representation $\mathbf{z}_{v} \in \{0,1\}^{\beta}$ using $\mathbf{r}_{v}$ and $\mathcal{N}^{k}(v)$. 
(iii). Uses  $\mathbf{z}_{v}$ to predict node labels and links in the graph. 
We proceed to describe each of these steps in detail.

\begin{figure}[h!]
\vspace*{-0.3cm}
    \centering
    \includegraphics[width=0.6\linewidth]{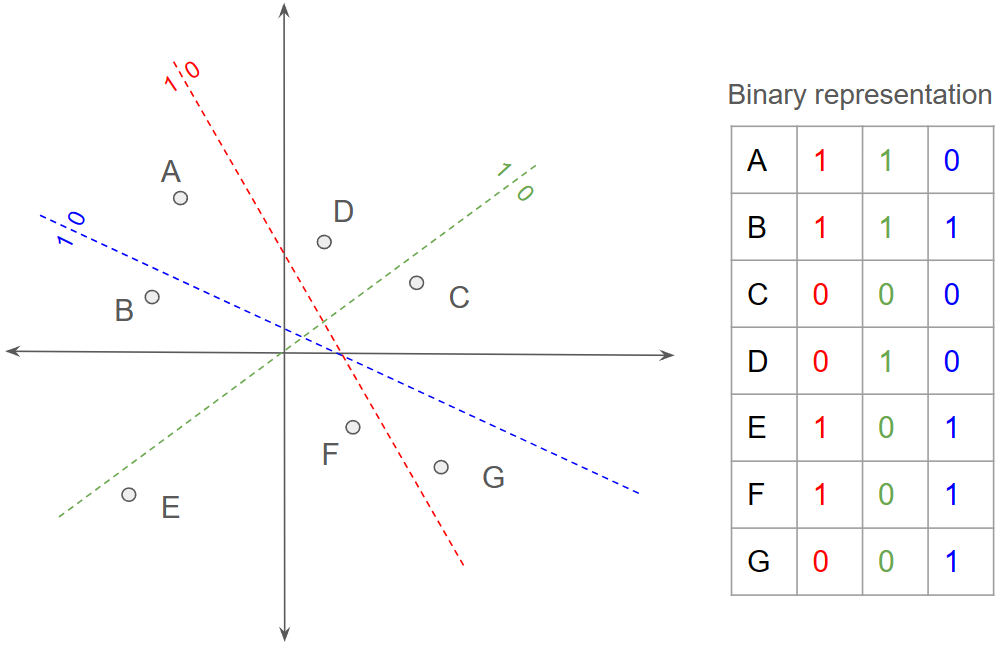}
    \vspace*{-12pt}
    \caption{Mapping node features to a binary space using Random Hyperplanes: points are assigned 1 or 0 based on their position relative to each hyperplane.}
    \vspace*{-0.61cm}
    \Description{}
    \label{fig:lsh_hyperplane}
\end{figure}

\subsection{Mapping Node Features to HD-space} \label{section: RHPT}

The first crucial step is to map node features from $\mathbb{R}^{d}$ space to $\{0,1\}^{\beta}$ HD space such that $\beta >> d$. We do this using random hyperplane tessellations \cite{Dirksensharp_randomHyperPlane}, a type of Locality Sensitive Hashing \cite{IndykLSH, cahrikar}. The key idea is to generate $\beta$ random hyperplanes in  $\mathbb{R}^{d}$ and set the corresponding co-ordinate of the HD representation of a node based on whether its feature vector in $\mathbb{R}^{d}$ lies on the positive or the negative side of the hyperplane. With  $\beta$ such hyper-planes in hand, we can generate  an $n$ dimensional binary sketch for each node. More precisely, let $\mathbf{Q}: \mathbb{R}^{d} \rightarrow \mathbb{R}^{\beta}$ be a matrix with each row $\mathbf{q}_{1}^{T}, \cdots \mathbf{q}_{\beta}^{T}$ drawn from  $\mathcal{N}(0,I_{d})$ and $\gamma$ be uniformly distributed on $[-\lambda, \lambda]$. For conciseness, we represent $\beta$ samples of $\gamma$ as $\Gamma \sim [-\lambda, \lambda]^{\beta}$. Using $\mathbf{Q}$ and $\Gamma$, a randomized binary sketch $\mathbf{r}_{v}$ for a node $v \in \mathcal{V}$ is given by $\mathbf{r}_{v} = \text{sign}(\mathbf{Q}\mathbf{x}_v + \Gamma)$. The offset $\Gamma$ when it is non-zero ensures that the sampled hyperplane does not pass through the origin. Note that the resulting binary sketch of each node is in fact a $\beta$-dimensional binary HD vector. Nodes with similar features will have similar HD representations; and nodes with dissimilar features will have dissimilar HD representations. 

\subsection{Neighborhood Aggregation Scheme}\label{section:HDGL_agg_scheme}

To simplify the mapping of nodes to their HD encodings, we assume that  nodes that have identical multi-sets of features over their $k$-hop neighborhoods are likely to also have identical local topologies. Under this assumption,  we can approximate  injectivity of the embedding $\mathbf{z}_v$ of each node $v$ to its HD representation by aggregating features of nodes over a $k$-hop neighborhood of $v$ as follows: 
$
\mathbf{z}_{v} = \phi \Big( \mathbf{r}_{v}, \varphi_{1}(\{\mathbf{r}_{j}; j \in \mathcal{N}^{1}(v) \}) , \ldots , \varphi_{K}(\{\mathbf{r}_{m}; m \in \mathcal{N}^{K}(v) \}) \Big)
$
where $\mathbf{z}_{v}$ denotes the HD embedding of node $v$ and $\phi(\cdot)$ and $\varphi_{k}(\cdot); k \in \{1,\cdots, K\}$  are injective functions.

\noindent\textbf{Ensuring that similar nodes have similar HD representations.} Suppose for a moment that the Binding operation  $\otimes$ is used to realize $\varphi_{k}(\cdot)$. Because binding of two similar vectors results in dissimilar vectors, it would adversely impact the utility of the HD representation for predicting node labels based on node features and local node topologies.

In contrast, we use Bundling operation because it is permutation-invariant and is well-suited for implementing the functions $\varphi_{k}(\cdot)$ under the assumption that nodes that have identical multi-sets of features over their $k$-hop neighborhoods are likely to also have identical local topologies. More precisely,
$
  \varphi_{k} \Big( \{\mathbf{r}_{j}; j \in \mathcal{N}^{k}(v) \} \Big) =  \mathop{\oplus}_{j \in \mathcal{N}^{k}(v)} \mathbf{r}_{j} 
$
where $\mathop{\oplus}_{j \in \mathcal{N}^{k}(v)} \mathbf{r}_{j} $ denotes bundling the HD representations of all of the $k$-hop neighbors of $v$. This operation is analogous to finding a high-dimensional vector that is most similar to all of the input high-dimensional vectors, similar to finding the mean in Euclidean spaces and like the mean, it is not necessarily injective. However, as noted by \cite{GIN_cite}, GNNs that use the mean to aggregate features over node neighborhoods  \cite{kipf2016semi,hamilton2017inductive} perform quite well in practice because the mean operation becomes increasingly injective when graphs consist of diverse node features, which is typically the case in real-world graphs.

\noindent\textbf{Binding for linking node with its neighbors.} Having chosen  Bundling  to realize $\varphi_{k}(\cdot)$, we proceed to consider how to realize $\phi(\cdot)$, which has to couple the representation of a node obtained using bundling (see above) with the representations of its neighbors.

As we will see shortly, if we choose to realize $\phi(\cdot)$ using the  Binding operation  $\otimes$, we obtain a representation that strikes a reasonable balance between discriminating between dissimilar graphs and generalizing over similar graphs.  Let $v \in \mathcal{V}$ and $w \in \mathcal{V}$ be two nodes in a graph. Suppose $\phi(\cdot)$ is realized using the Binding Operation $\otimes$ and $\psi_{k}(\cdot)$ is realized using the Bundling Operation. Considering only 1-hop and 2-hop neighbors of $v$, the latent representation of $v$ and $w$ will be 
\begin{equation}\label{eq.example}
\begin{split}
\mathbf{z}_{v} &=  \mathbf{r}_{v}  \otimes  \Big( \mathop{\oplus}_{j \in \mathcal{N}^{1}(v)}  \mathbf{r}_{j} \Big)  \otimes  \Big( \mathop{\oplus}_{m \in \mathcal{N}^{2}(v)}  \mathbf{r}_{m}  \Big) \\
\mathbf{z}_{w} &=   \mathbf{r}_{w}  \otimes  \Big( \mathop{\oplus}_{j \in \mathcal{N}^{1}(w)}  \mathbf{r}_{j}  \Big)  \otimes  \Big( \mathop{\oplus}_{m \in \mathcal{N}^{2}(w)} \mathbf{r}_{m}  \Big)
\end{split}
\end{equation}
Suppose we have 
\begin{equation*}
    \begin{split}
        \mathbf{r}_{w} \approx \mathbf{r}_{v}; \quad \; \Big( \mathop{\oplus}_{j \in \mathcal{N}^{1}(v)}  \mathbf{r}_{j} \Big)  \approx \Big( \mathop{\oplus}_{j \in \mathcal{N}^{1}(w)} \mathbf{r}_{j} \Big) \\
    \end{split}
\end{equation*}

\noindent and  $\displaystyle{\mathop{\oplus}_{j \in \mathcal{N}^{2}(v)} \mathbf{r}_{j} }$ is dissimilar to $ \displaystyle{\mathop{\oplus}_{j \in \mathcal{N}^{2}(w)} \mathbf{r}_{j}}$. Then $\mathbf{z}_{v}$ and $\mathbf{z}_{w}$ will be dissimilar since binding is reflecting i.e

$$d_{H}\Big[ \Big( \mathop{\oplus}_{j \in \mathcal{N}^{2}(v)}  \mathbf{r}_{j} \Big) , \Big( \mathop{\oplus}_{j \in \mathcal{N}^{2}(w)}  \mathbf{r}_{j}  \Big) \Big] \approx d_{H} \Big[ \mathbf{z}_{v} , \mathbf{z}_{w}   \Big]$$

It is easy to see that if all the terms in $\mathbf{z}_{v}$ and $\mathbf{z}_{w}$ are similar, i.e., $v$ and $w$ have similar node features, and similar 1-hop and 2-hop neighborhoods, then $\mathbf{z}_{v} \approx \mathbf{z}_{w}$ as desired.
On the other hand, if the encodings of the features of the 2-hop neighbors of both $v$ and $w$ differ,  $\mathbf{z}_{v}$ and $\mathbf{z}_{w}$ will be dissimilar as desired. When not only the 2-hop neighbors but also the 1-hop neighbors of both $v$ and $w$ are dissimilar, the dissimilarity between $\mathbf{z}_{v}$ and $\mathbf{z}_{w}$ becomes even greater. The key property that makes binding attractive for realizing the function $\phi(\cdot)$ is its reflectivity, i.e., the distance between two HD vectors remains unchanged when both of them are bound to the same third vector.

On the other hand, it is easy to see that Bundling is not a good choice for realizing $\phi(\cdot)$. Because bundling is a bitwise majority operation, bundling the node vector $\mathbf{r}_{v}$ with neighborhood information vector $\oplus_{j \in \mathcal{N}^{k}(v)} \mathbf{r}_{j}$; results in a vector where the information from a node's neighborhood suppresses the information about the node's features, which is an undesirable property.

\noindent\textbf{Improving injectivity via Rotation. } Because Binding is associative and commutative, it can adversely impact the  injectivity of $\varphi_{k}(\cdot)$ and hence the representational power of our HD model. For example, suppose $v \in \mathcal{V}$ and $w \in \mathcal{V}$ are two nodes in a graph with encoding obtained using  HD mapping that uses Bundling to realize $\varphi_k$ and binding to realize $\phi(\cdot)$ and $\mathbf{z}_{v}$ and $\mathbf{z}_{w}$ are computed as per Eq. \eqref{eq.example}. Consider the scenario in which $\mathbf{x}_{v} = \mathbf{x}_{w}$,   $\mathcal{N}^{1}(v) = \mathcal{N}^{2}(w)$ and $\mathcal{N}^{2}(v) = \mathcal{N}^{1}(w)$. Now, assuming $\mathcal{N}^{1}(v) \neq \mathcal{N}^{2}(v)$, the representation of $w$ can be rewritten as 
$
\mathbf{z}_{w} =   \mathbf{r}_{v} \otimes  \Big( \mathop{\oplus}_{j \in \mathcal{N}^{2}(v)}  \mathbf{r}_{j} \Big)  \otimes  \Big( \mathop{\oplus}_{m \in \mathcal{N}^{1}(v)}  \mathbf{r}_{m}  \Big)
$.
Since, Binding is both associative and commutative, the latent representations $\mathbf{z}_{v}$ and $\mathbf{z}_{w}$ become equal, despite their 1-hop and 2-hop neighbors being different. This adversely impacts the representation's ability to discriminate between dissimilar graphs. To address this issue, we utilize the Rotation $\Pi$ operator. Now, the latent representation for nodes $v \in \mathcal{V}$ and $w \in \mathcal{V}$ changes:
\begin{equation*}
\begin{split}
\mathbf{z}_{v} =  \mathbf{r}_{v} \otimes  \Pi \Big( \mathop{\oplus}_{j \in \mathcal{N}^{1}(v)} \mathbf{r}_{j}  \Big)  \otimes   \Pi \; \Pi \Big( \mathop{\oplus}_{m \in \mathcal{N}^{2}(v)} \mathbf{r}_{m} \Big) \\
\mathbf{z}_{w} =   \mathbf{r}_{v}  \otimes \Pi \Big( \mathop{\oplus}_{j \in \mathcal{N}^{2}(v)} \mathbf{r}_{j} \Big)  \otimes \Pi \; \Pi \Big( \mathop{\oplus}_{m \in \mathcal{N}^{1}(v)} \mathbf{r}_{m} \Big)
\end{split}
\end{equation*}

\noindent where $\Pi(\cdot)$ and $\Pi \; \Pi(\cdot)$ denote single and double rotation operations respectively. Because the rotation of HD vector yields a HD vector that is dissimilar or orthogonal to the original, now $\mathbf{z}_{v} $ and $ \mathbf{z}_{w}$ will be dissimilar. That is,
\begin{equation*}
    \begin{split}
       d_{H} \Big[ \Pi \Big( \mathop{\oplus}_{j \in \mathcal{N}^{1}(v)} \mathbf{r}_{j} \Big), \Pi \; \Pi \Big( \mathop{\oplus}_{m \in \mathcal{N}^{2}(v)} \mathbf{r}_{m} \Big) \Big] \approx 0.5 \\
        d_{H} \Big[ \Pi \Big( \mathop{\oplus}_{j \in \mathcal{N}^{2}(v)}  \mathbf{r}_{j}  \big), \Pi \; \Pi \Big( \mathop{\oplus}_{m \in \mathcal{N}^{1}(v)}  \mathbf{r}_{m}  \Big) \Big] \approx 0.5 \\
    \end{split}
\end{equation*}

Hence, considering only 1-hop and 2-hop neighbors, the proposed  aggregation scheme to be used by HDGL for computing the HD representation of a node $v \in \mathcal{V}$ in the graph is given by

\begin{equation}
\label{HD_GL_agg_scheme}
\begin{split}
\mathbf{z}_{v} =  \mathbf{r}_{v} \otimes  \Pi \Big( \mathop{\oplus}_{j \in \mathcal{N}^{1}(v)} \mathbf{r}_{j}  \Big)  \otimes   \Pi \; \Pi \Big( \mathop{\oplus}_{m \in \mathcal{N}^{2}(v)} \mathbf{r}_{m} \Big) \\
\end{split}
\end{equation}

\begin{algorithm}[ht!]
\small
\caption{Pseudo-code for HDGL}\label{alg:cap}
\SetAlgoNlRelativeSize{-2}

\KwIn{$\mathcal{G} = (\mathcal{V}, \mathcal{E})$, Node Features $\mathbf{x}_{v}$, Train Nodes $\mathcal{T}_{\ell} \subset \mathcal{V}$ for each class $\forall \ell \in \{1, \cdots, L \}$ and Unlabeled Test Nodes $\mathcal{V}_{\text{test}}$}

Sample $\mathbf{Q}:[\mathbf{q}_{1}^{T} \sim \mathcal{N}(0,I_{d}), \cdots ,\mathbf{q}_{\beta}^{T} \sim \mathcal{N}(0,I_{d})]$\;
Sample $\Gamma \sim [-\lambda, \lambda]^{\beta}$\;

\For{$v \in \mathcal{V}$}{
    $\mathbf{r}_{v} \gets \text{sign}(\mathbf{Q}\mathbf{x}_v + \Gamma)$\;
}

\For{$v \in \mathcal{V}$}{
    $\mathbf{z}_{v} \gets \mathbf{r}_{v} \otimes  \Pi \Big( \mathop{\oplus}_{j \in \mathcal{N}^{1}(v)} \mathbf{r}_{j}  \Big)  \otimes   \Pi \; \Pi \Big( \mathop{\oplus}_{m \in \mathcal{N}^{2}(v)} \mathbf{r}_{m} \Big)$\;
}

\If{Node Label Prediction Task}{
    
    \For{$\ell \in \{1, \cdots, L \}$}{
        $\mathbf{c}_{\ell} \gets \mathop{\oplus}_{i  \in \mathcal{T}_{\ell}}  \mathbf{z}_{i}; \quad$  \#Computing Label Hyper Vectors\;
    }
    
    \For{$u \in \mathcal{V}_{\text{test}}$}{
        $\hat{y}_{u} \gets \mathop{\argmin}_{\ell \in \{1, \cdots, L \}} d_{H}[\mathbf{z}_{u}, \mathbf{c}_{\ell}]; \: $ \#Test-set Inference\;
    }
}

\If{Link Prediction Task}{
    $\mathbf{e}^{+} \gets \mathop{\oplus}_{(u,v) \in \mathcal{E}}(\mathbf{z}_{u} \otimes \mathbf{z}_{v})$ and $ \mathbf{e}^{-} \gets \mathop{\oplus}_{(u,v) \notin \mathcal{E}}(\mathbf{z}_{u} \otimes \mathbf{z}_{v})$\;
    Compute $\mathbf{Z}^{+}$ and $\mathbf{Z}^{-}$ using Eqn. \eqref{eq:Zplusminus}\;
    $\mathbf{D}^{+} \gets dist(\mathbf{Z}^{+}, \mathbf{Z})$ and $\mathbf{D}^{-} \gets dist(\mathbf{Z}^{-}, \mathbf{Z})$\;
    \For{$i = 1$ \textbf{to} $N$ }{
    \For{$j = 1$ \textbf{to} $N$}{
        \If{$D^+_{ij} < D^-_{ij}$}{
            $\hat{A}_{ij} \gets \sigma \big((1 - D^+_{ij}) + D^-_{ij}\big)$\;
        }
        \Else{
            $\hat{A}_{ij} \gets \sigma \big(D^+_{ij} - (1 - D^-_{ij})\big)$\;
        }
    }}
}

\end{algorithm}

\subsection{Node Label Prediction using HDGL}

With the procedure for constructing a HD representation of nodes and their local topologies in place, we turn to the task of using the resulting HD node representation to predict the node labels using $\mathbf{z}_{v}$. Here, we follow the standard approach to predicting class labels of HD encoded objects \cite{ge2020classification}. Specifically, during the learning phase, we Bundle the collection of HD encodings of nodes belonging to each class  yielding   a HD Class Hyper-Vector for each of the classes.  During the inference phase, given the HD vectors for each of the classes, we predict the label of an unlabeled node $u$ as simply the label associated with the class hyper-vector that is, among all class HD class vectors is closest to the HD vector $\mathbf{z}_u$, the encoding of $u$. More precisely let $\forall \ell \in \{1, \cdots, L \}$,  $\mathcal{T}_{\ell} \subset \mathcal{V}$ be the subset of nodes that make up the training set with label $\ell$ and $\mathcal{V}_{\text{test}} \subset \mathcal{V}$ be the subset of unlabeled nodes in the graph. The class HD vectors are constructed for each class as follows:
$
   \mathbf{c}_{\ell} = \mathop{\oplus}_{i  \in \mathcal{T}_{\ell}}  \mathbf{z}_{i} \quad ;\; \forall \ell \in \{1, \cdots, L \}
$, resulting in  a set of class HD vectors $\{\mathbf{c}_{1}, \cdots, \mathbf{c}_{L} \}$. Let $u \in \mathcal{V}_{\text{test}}$ be a node for which the class label is to be predicted. We predict the class label for $u$ using its HD representation and the HD representations of each of the classes as follows:
$
   \hat{y}_{u} = \mathop{\argmin}_{\ell \in \{1, \cdots, L \}} d_{H}[\mathbf{z}_{u}, \mathbf{c}_{\ell}]
$.

\subsection{Link Prediction Using HDGL}

Our approach takes advantage of a technique introduced by \citet{KanervaSDM, HD_computing_intro}'s for representing and retrieving data using HD representation. Lets consider a graph with 7 nodes and there are 3 edges in the graph betweem node $(1,2)$, $(3,6)$ and $(5,7)$. By binding latent representation between pairs of nodes which have edges between them and bundling the resulting vectors, we obtain a sum-vector that encapsulates the edge information of the graph: $\mathbf{e}^{+} = (\mathbf{z}_{1} \otimes \mathbf{z}_{2}) \oplus (\mathbf{z}_{3} \otimes \mathbf{z}_{6}) \oplus (\mathbf{z}_{5}\otimes \mathbf{z}_{7})$.

\noindent Edge information can be recovered solely using hypervectors $\mathbf{e}^{+}$ and $\mathbf{z}_{i}; \forall i  \in \mathcal{V}$. For eg. $ \mathbf{z}_{2} \approx \mathbf{e}^{+} \otimes \mathbf{z}_{1}$ and $ \mathbf{z}_{7} \approx \mathbf{e}^{+} \otimes \mathbf{z}_{5}$.

For the transductive link prediction problem, we leverage on the assumption that the underlying semantics governing existing links, which are based on latent representations of nodes in an edge, will also apply to new edges. For instance, if nodes $(1,2)$ are connected and we are predicting edges involving node $4$ and if $\mathbf{z}_{1} \approx \mathbf{z}_{4}$, then we would expect an edge $(4,2)$ to be predicted. Note that it's also crucial to accurately predict non-edges to minimize false positives. Therefore, we also make use of hypervector $\mathbf{e}^{-}$ to represent information about edges that are \emph{absent} in the graph.

Following this overview, we can now delve into a detailed explanation of our model. Formally:
$
         \mathbf{e}^{+} = \mathop{\oplus}_{(u,v) \in \mathcal{E}}(\mathbf{z}_{u} \otimes \mathbf{z}_{v}); \quad \mathbf{e}^{-} = \mathop{\oplus}_{(u,v) \notin \mathcal{E}}(\mathbf{z}_{u} \otimes \mathbf{z}_{v}).
$ Given this encoding of edges that are present and the edges that are absent in the graph, we can proceed to predict links as follows:

To find putative new links we Bind $\otimes$ the edge information hypervector $\mathbf{e}^{+}$ and the latent representation of all the nodes in the graph. This results in a new matrix $\mathbf{Z}^{+}$ where the $i^{th}$ row is $\mathbf{z}_{i} \otimes \mathbf{e}^{+}$; representing the possible node representation of the target node in a link, with the corresponding source node representation being $\mathbf{z}_{i}$. We follow the same procedure for the  hypervector encoding information about the edges that are absent, resulting in $\mathbf{Z}^{-}$. Precisely:
\vspace*{-8pt}
\begin{equation}
\begin{array}{cc}
\mathbf{Z}^{+} = 
\begin{bmatrix}
\mathbf{z}_{1} \otimes \mathbf{e}^{+} \\
\vdots \\
\mathbf{z}_{N} \otimes \mathbf{e}^{+}
\end{bmatrix} &
\mathbf{Z}^{-} = 
\begin{bmatrix}
\mathbf{z}_{1} \otimes \mathbf{e}^{-} \\
\vdots \\
\mathbf{z}_{N} \otimes \mathbf{e}^{-}
\end{bmatrix}
\end{array}
\label{eq:Zplusminus}
\end{equation}

Now, we find pairwise normalized distances between rows of $\mathbf{Z}^{+}$ and rows of $\mathbf{Z}$. This results in $\mathbf{D}^{+} = dist(\mathbf{Z}^{+}, \mathbf{Z}) \in [0,1]^{N \times N}$ , a  matrix representing distances. If entry $(i,j)$ is close to zero, it indicates that the "distance" between row $i$ of $\mathbf{Z}^{+}$ is close to row $j$ in $\mathbf{Z}$, suggesting the possibility of a link $i$ and $j$. Similarly, we find pairwise normalized distances between rows of $\mathbf{Z}^{-}$ and rows of $\mathbf{Z}$ yielding  $\mathbf{D}^{-} = dist(\mathbf{Z}^{-}, \mathbf{Z})$. An entry close to zero in $\mathbf{D}^{-}$ suggests the absence of an edge between the corresponding nodes.

We now generate a prediction for the adjacency matrix, denoted by $\hat{\mathbf{A}}$, where $\hat{A}_{ij} \in [0,1]$. We predict values that fall within this interval because the evaluation metrics used are AUC-ROC and Average Precision. For real-world applications, an appropriate threshold can be selected depending on the specific use case to effectively interpret these predictions. To obtain the final predicted adjacency matrix of probabilities, which reflects the likelihood of an edge being present based on $\mathbf{D}^{+}$ and $\mathbf{D}^{-}$, we perform the following:

\vspace*{-5pt}
    \[
\hat{A}^{}_{ij} =
\begin{cases}
    \sigma \big((1 - D^{+}_{ij}) + D^{-}_{ij}\big) & \text{if } D^{+}_{ij} < D^{-}_{ij} \\
    \sigma \big(D^{+}_{ij} - (1-D^{-}_{ij})\big) & \text{else} 
\end{cases}
\]

\noindent where $\sigma(\cdot)$ is the sigmoid function. The rationale behind this operation lies in the comparison between $D^{+}_{ij}$ and $D^{-}_{ij}$. When $D^{+}_{ij} < D^{-}_{ij}$, it indicates a higher likelihood of a connection between nodes $i$ and $j$ according to the distance metric. Hence, we apply the operation $\sigma((1 - D^{+}_{ij}) + D^{-}_{ij})$ to obtain a value closer to one. We use the same logic when $D^{+}_{ij} \geq D^{-}_{ij}$ to obtain a value close to zero.

\section{Experiments}
\label{sec:exp}

We proceed to report results of extensive experiments comparing HDGL with  a set of state-of-the-art node classification methods and link prediction methods, including several GNN models, in terms of both predictive performance and runtime. The code for HDGL can be found at: {\textcolor{blue}{\url{https://github.com/Abhishek-Dalvi410/HDGL}}}.

\subsection{Data sets}

Since HDGL is designed to ensure that nodes with similar features and neighborhoods have comparable representations—often resulting in identical node labels—we focus exclusively on graphs with these characteristics as benchmark datasets for this study.

For our experiments, we utilize commonly used node labeling graph benchmarks: the CORA graph dataset \cite{planetoid_yang}, sourced from \citet{mccallum2000automating}; the CiteSeer graph dataset \cite{planetoid_yang}, sourced from \citet{10.1145/276675.276685}; and the PubMed dataset \cite{Pubmed_ref}. Additionally, we include three more datasets from the Microsoft Academic Graph: Coauthor CS and Coauthor Physics from \citet{shchur2018pitfalls}, and the BlogCatalog Graph dataset from \citet{dataset_blogcatalog}. Additionally, we introduce a new version of the DBLP dataset, built upon the works of \citet{Pan2016TriPartyDN_DBLP}. We remove nodes with no neighbors in the graph obtained from \citet{Pan2016TriPartyDN_DBLP}. The original node features from \citet{Pan2016TriPartyDN_DBLP} consist of manuscript titles. We utilize BERT \cite{Devlin2019BERTPO} to obtain embeddings for the titles and employ them as node features. The data sets and their statistics are given in Table \ref{tab:dataset_statistics}. 

We adhere to the Train/Validation/Test splits described in \citet{planetoid_yang}, \citet{kipf2016semi}, and \citet{veličković2018graph} for the CORA, CiteSeer, and PubMed datasets, respectively. For the Coauthor CS, Coauthor Physics, BlogCatalog, and DBLP datasets, we employ the data splitting strategy outlined in \citet{shchur2018pitfalls}.

We test transductive link prediction models on CORA, CiteSeer, and Pubmed datasets and with train/val/test splits from \citet{kipf2016variational_VGAE} work i.e validation and test sets each contain 5\% and 10\% of links.

\begin{table}[ht!]
  \centering
  \vspace*{-7pt}
  \caption{Dataset Summary.} 
  \vspace*{-10pt}
    \resizebox{0.43\textwidth}{!}{%
  \begin{tabular}{lccccc}
    \toprule
    Dataset & Nodes & Edges & Features & Feature Type \\
    \midrule
    CORA & 2708 & 5429 & 1433 & $\{0,1\}$  \\
    CiteSeer & 3327 & 4732 & 3703 & $\{0,1\}$ \\
    PubMed & 19717 & 44338 & 500 & $(0,1)$ \\
    BlogCatalog & 5196 & 171743 & 8189 & $\mathbb{Z}^{+}$ \\
    Coauthor CS & 18333 & 81894 & 6805 & $\mathbb{Z}^{+}$  \\
    Coauthor Physics & 34493 & 247962 & 8415 & $\mathbb{Z}^{+}$\\
    
    DBLP & 17725 & 105781 & 768 & \hspace{-0.4em}$\mathbb{R}$ \\
    \bottomrule
  \end{tabular}
  }
   \label{tab:dataset_statistics}
   \vspace*{-15pt}
\end{table}

\subsection{Models}

For semi-supervised node classifcation task, we compare the performance of HDGL with that of  several  strong baselines Logistic Regression (LogReg), DeepWalk \cite{deepwalk_perozzi} and Label Propagation \cite{Zhu2002LearningFL_labelprop}. We also directly compare HDGL with state-of-the-art models: Graph Convolutional Network (GCN) \cite{kipf2016semi}, Graph Attention Network (GAT) \cite{veličković2018graph}, and lastly, a fast GNN model, namely, Simplified Graph Convolutional Network (SGC) \cite{Wu2019SimplifyingGC} which is essentially a logistic regression model trained using features extracted from $k$-hop neighborhoods of nodes as input. We also assess our model against RelHD \cite{Tajana_RelHD}, an existing HD algorithm used for node classification

For the transductive link prediction task, we evaluate our model against standard link prediction baselines:- DeepWalk \cite{deepwalk_perozzi} and Spectral clustering (SC) \cite{Tang2011LeveragingSM_spectral_clustering}. We also compare our model against graph autoencoder models:- Graph Autoencoder (GAE) and Variational Graph Autoencoder (VGAE) from \citet{kipf2016variational_VGAE}.

\subsection{HDGL Model Specifications}

To ensure that HDGL are directly comparable with most GNN architectures which typically use 2 layers, and hence aggregate information over 2-hop neighbors, we limit HDGL to aggregate information over at most 2-hops i.e HD node representations are calculated as per Eq.\eqref{HD_GL_agg_scheme}.

Because Bundling of an even number of HD vectors can result in ties that would need to be broken using tie-breaking policy (See Section \ref{HD-comp_background}), we opt to randomly sample odd number of neighbors, specifically, 11 of the 1-hop neighbors and 21 of the 2-hop neighbors during the construction of HD node representations by HDGL for datasets except BlogCatalog. For BlogCatalog we sample 21 of the 1-hop neighbors and 51 of the 2-hop neighbors since Blogcatalog graph is denser than the other datasets. 

For node classification in CORA, CiteSeer, and BlogCatalog datasets, we map node features to $50,000$-dimensional HD vectors through random projections. Similarly, for the PubMed, Coauthor CS, Coauthor Physics, and DBLP datasets, we map node features to $20,000$-dimensional HD vectors using the same method. While these choices may seem arbitrary, it's crucial to note that selecting dimensions between $20,000$ and $50,000$ typically suffices to maintain the desired properties of HD-computing. This assertion finds extensive support in the research from \citet{KanervaSDM, HD_computing_intro}.

\begin{table*}[ht!]
\begin{center}
\caption{Average Test Set Accuracy and Standard Deviation (in Percentage) Over 10 Random Weight Initializations for Deep Learning Models and 10 Randomized Instantiations of RelHD and HDGL Across  Datasets (* denotes datasets with $\{0,1\}$ features. In these datasets, RelHD performs comparably to HDGL)}
\vspace*{-0.3cm}
\begin{tabular}{ |c|| c| c| c| c| c| c| c| }
\hline
\textbf{Method} & \textbf{Cora}$^{*}$ & \textbf{Citeseer}$^{*}$ & \textbf{Pubmed} & \textbf{BlogCat} & \textbf{CompSci}  & \textbf{Physics} & \textbf{DBLP} \\
\hline
\hline
LogReg  & 52.2 $\pm$ 0.5 & 46.3 $\pm$ 0.4 & 67.9 $\pm$ 0.5 & 65.3 $\pm$ 1.0 & 86.8 $\pm$ 0.6 & 88.2 $\pm$ 0.9 & 56.2 $\pm$ 3.0 \\

DeepWalk  & 65.3 $\pm$ 1.2 & 46.2 $\pm$ 0.9 & 69.3 $\pm$ 0.9 & 58.6 $\pm$ 1.3 & 78.1 $\pm$ 0.9 & 86.4 $\pm$ 0.9 & 66.1 $\pm$ 2.2 \\

Label Prop  & 70.9 $\pm$ 0.0 & 47.7 $\pm$ 0.0  & 71.3 $\pm$ 0.0  & 47.8 $\pm$ 3.8 & 76.6 $\pm$ 0.9  & 85.6 $\pm$ 1.1 & 66.4 $\pm$ 1.6 \\
\hline
GCN  & 81.9 $\pm$ 0.6  & 70.3 $\pm$ 0.5 & 79.0 $\pm$ 0.4 & 69.5 $\pm$ 0.4 & 91.0 $\pm$ 0.7 &  92.3 $\pm$ 0.6 & 72.1 $\pm$ 0.8 \\

GAT  & 82.8 $\pm$ 0.5  & 72.0 $\pm$ 0.6 & 79.1 $\pm$ 0.3 & 65.3  $\pm$ 1.2 & 90.9 $\pm$ 0.6  &  92.5 $\pm$ 0.9 & 75.4 $\pm$ 1.0\\

SGC  & 79.8 $\pm$ 0.5  & 68.5 $\pm$ 0.4 & 75.4 $\pm$ 0.5 & 70.8 $\pm$ 0.3 & 89.1 $\pm$ 0.6 &  92.1 $\pm$ 0.7 & 76.3 $\pm$ 0.9 \\
\hline
RelHD & 78.9 $\pm$ 0.5  & 69.3 $\pm$ 0.4 & 69.1 $\pm$ 0.4 & 60.9 $\pm$ 1.2 &  85.5 $\pm$ 0.8 & 85.3 $\pm$ 0.9 & 44.9  $\pm$ 0.2 \\

HDGL & 79.5 $\pm$ 0.7  & 70.0 $\pm$ 0.6 & 76.8 $\pm$ 1.1 & 69.3 $\pm$ 0.5  &  88.9 $\pm$ 0.4 & 91.0 $\pm$ 0.9 & 68.7 $\pm$ 1.2 \\
\hline
\end{tabular}
\vspace*{-0.4cm}
\label{tab:acc_results}
\end{center}
\end{table*}

\subsection{Training and Evaluation}

Iterative machine learning methods like GNN use separate validation data to stop training and tune hyperparameters, alongside training data. In contrast, methods such as HDGL which involve no iterative training do not need validation data. Hence, to ensure fair comparison, HDGL uses training and validation sets together for learning, whereas GNN use training data for training the model, validation data for early stopping and hyperparameter tuning.

In the case of GNN methods that rely on iterative training, we choose the models with hyperparameters tuned to obtain optimal performance on validation data. We used grid search over 12 different configurations for tuning these hyperparameters. For GCN and GAT, we optimized the following hyperparameters: the number of hidden units in the first layer, learning rate, and weight decay. Specifically, GCN models were tested with 32 and 64 hidden units, while GAT models were tested with 4 and 8 attention heads, each having 8 hidden units. Both GCN and GAT models were evaluated with weight decays of 1e-3, 1e-4, and 5e-4. SGC, which does not have hidden layers, had its hyperparameters tuned for learning rate and weight decay. The weight decay tuning of SGC was over the following values 1e-3, 5e-4, 1e-4, 5e-5, 1e-5, and 5e-6.

GCN, GAT, and SGC models were tuned with learning rates of 1e-2 and 1e-3, using early stopping criteria set to 25 epochs for the 1e-2 learning rate and 50 epochs for the 1e-3 learning rate.

As LogReg, DeepWalk, and Label Prop represent well-established baselines, we do not conduct a hyperparameter search. For Label Propagation, we use 20 iterations of propagation (except for blogcatalog dataset where we use 10) and set the alpha factor \footnote{Alpha factor in Label Propa determines the balance between existing and propagated labels. Higher alpha prioritizes existing labels; lower alpha favors propagated ones.} to 0.5. For DeepWalk, we utilize the DGL library, setting the embedding dimension to 32 and the number of walks to 20, with a batch size of 64. We sample such walks for all nodes in the graph and train DeepWalk model for 10 epochs and then employ logistic regression for classification, incorporating early stopping based on the validation set. 

Following \citet{Tajana_RelHD}, the RelHD algorithm was employed with 10,000 dimensional random basis HD vectors, as the performance of RelHD plateaus around this dimensionality.

We report measured performance on the test data, averaged across 10 random selections of pertinent parameters. These parameters include the random hyperplanes employed for constructing high-dimensional representations used by HDGL, random basis vector initializations for RelHD, weight initializations of the deep learning models, and data splits for datasets such as Coauthor Physics, Coauthor CS, BlogCatalog, and DBLP. For Cora, Citeseer, and Pubmed, we use predefined splits given in \citet{planetoid_yang}.

For the link prediction task, we rely on the results directly obtained from the paper for GVAE from \citet{kipf2016variational_VGAE}.

\begin{figure}[h!]
  \centering
  \vspace*{-5pt}
  \includegraphics[width=0.9\linewidth]{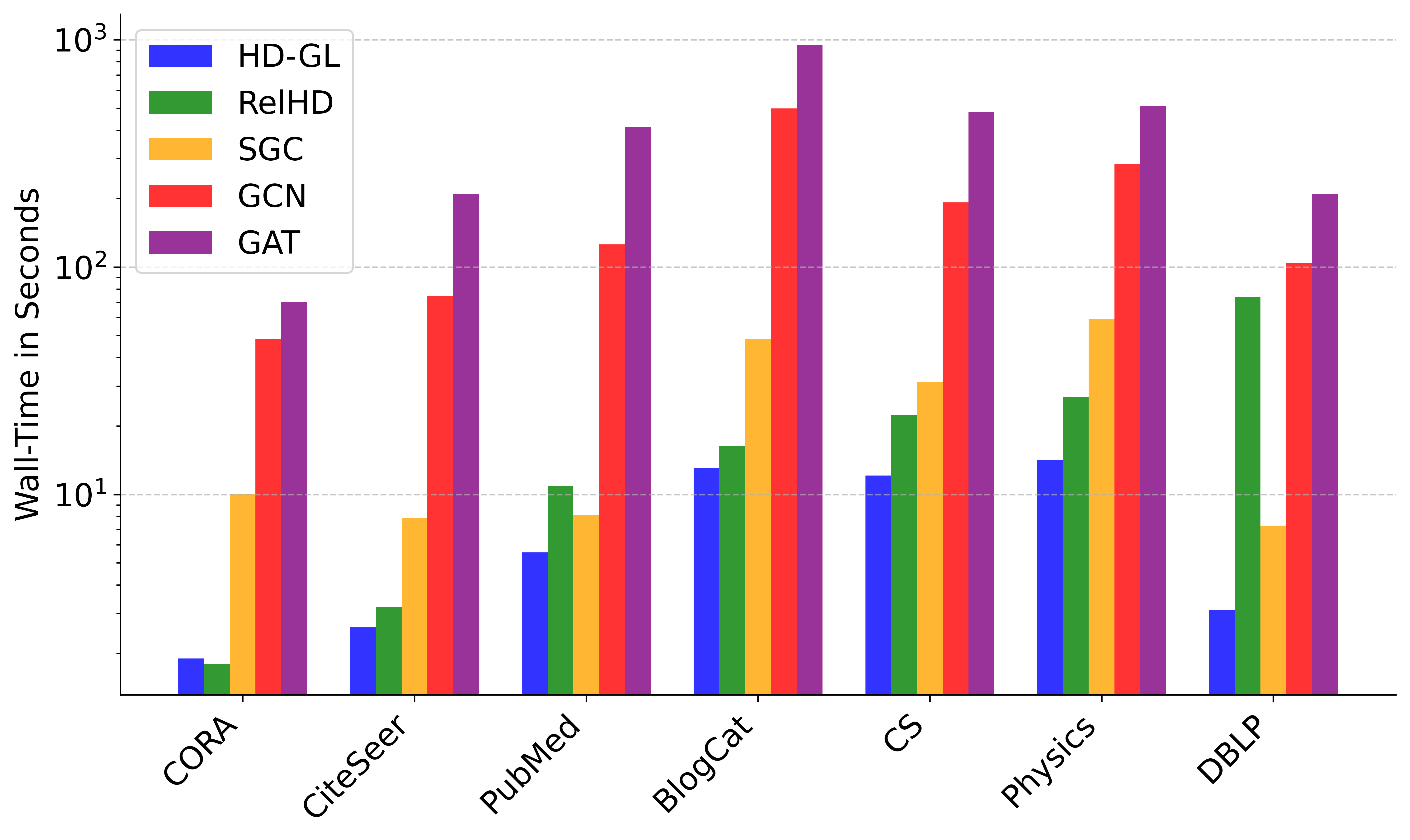}
  \vspace*{-0.4cm}
  \caption{Comparison of Learning Times (in seconds) i.e  Across Datasets for HDGL and RelHD with GCN, GAT and SGC with Hyperparameter Tuning with Exploration of Search Space with 12 configurations. Runtime for HDGL and RelHD exclusively includes only time Elapsed for learning label representations using train/validation data.}
  \vspace*{-15pt}
  \label{fig:Run_time_compare}
  \Description{}
\end{figure}

\subsection{Implementation Details}

All experiments were conducted on Google Colab Pro with High-memory CPU. To ensure fair comparison, although not ideal for HDGL, we used the PyTorch framework \cite{paszke2019pytorch}, complemented by the Deep Graph Library (DGL) \cite{Wang2019DGL} for importing datasets and implementing deep learning models. Due to the lack of support for bitwise majority operations in PyTorch, we convert our vectors to the bipolar space and utilize signed addition for compatibility. As mentioned in Section \ref{HD-comp_background}, HD computing works in both binary and bipolar spaces, allowing us to switch between them as necessary. None of our experiments used parallel execution or GPUs. We have also implemented PyTorch version of RelHD, as the experiments detailed in \citet{Tajana_RelHD} were reliant on hardware architecture implementations. We undertake this effort to compare with our model and use RelHD as a baseline in this study.

\begin{table*}[ht!]
\centering
\caption{Class Incremental Node Labeling Results. No hyperparameter grid search is used for GCN and SGC algorithms. Wall-time (in seconds) includes both training and inference of the model. HDGL does not require retraining from $t=2$.}
\vspace*{-0.3cm}
\label{tab:incremental_results}
\begin{tabular}{|c|c|c|c|c|c|c||c|c|c|c|}
\hline
\multirow{2}{*}{Model} & & \multicolumn{5}{c||}{BlogCatalog} &  \multicolumn{4}{c|}{Coauthor Physics}\\ \cline{3-11} 
 & & t = 1  & t = 2  & t = 3 &  t = 4 & t = 5 & t=1 & t=2 & t=3 &t=4 \\ \hline
\multirow{2}{*}{GCN} & Accuracy & 94.94 & 85.81 & 75.16 &  74.82 & 69.61 & 98.95 & 97.33 & 95.57 & 92.58 \\
 & Runtime & 51.2s & 18.7s & 16.5s &  33.9s & 23.9s & 65.9s & 62.3s & 65.8s & 51.2s\\
\hline
\multirow{2}{*}{SGC} & Accuracy & 94.56 & 82.31 & 73.23 &  72.72 & 69.81 & 98.94 & 97.39 & 95.36 & 91.81\\
 & Runtime & 6.3s & 3.72s & 3.43s & 3.73s & 3.6s & 9.1s & 8.5s & 8.8s & 9.7s\\
\hline
\multirow{2}{*}{HDGL} & Accuracy & 89.11 & 88.49 & 75.56 &  71.73 & 69.91 & 98.56 & 94.87 & 92.53 & 90.8\\
 & Runtime & 20.2s & 0.3s & 0.4s &  0.4s & 0.6s & 38.3s & 1.5s & 1.7s & 1.7s \\
\hline
\end{tabular}
\vspace*{-10pt}
\end{table*}

\subsection{Results}

\noindent\textbf{Node Classification Performance. }The results of our comparison of HDGL with the other methods are shown in Table \ref{tab:acc_results}. We observe that HDGL without the need for computationally expensive iterative training or hyperparameter optimization achieves node prediction accuracy that is competitive with those of state-of-the-art GNN models (GCN, GAT, SGC) across most of the benchmark data sets, in each case reaching performance that is within 1-2\% of the best performing model.  HDGL also consistently outperforming basic deep learning baseline methods (LogReg, DeepWalk, LabelProp). The notable advantage of HDGL over RelHD lies in its ability to accommodate features beyond the binary $\{0,1\}$ range, as evident from the results presented in Table \ref{tab:acc_results}, where HDGL significantly outperforms. However, when dealing with datasets featuring $\{0,1\}$ attributes like CORA and Citeseer, the performance of HDGL and RelHD appears nearly identical. 

\noindent\textbf{Node Classification Run-times. }
Figure \ref{fig:Run_time_compare} compares the run-time for the node classification task between HDGL and RelHD (both of which requires no hyperparameter tuning) with those of GCN, GAT, and SGC (with hyperparameter tuning), alongside the run-time of HDGL. We observe that, the run-time of HDGL is substantially less than that of all the state-of-the-art GNNs. While in most cases, RelHD exhibits slightly or significantly worse run times compared to HDGL, it's crucial to note that HDGL consistently delivers superior performance across various datasets when compared to RelHD.

It is important to note that although we used grid search over 12 configurations for hyperparameter optimization, typically deep learning models explore a much larger parameter space. This expanded search naturally comes with increased runtime, especially when GPUs are utilized, further widening the energy efficiency gap between Graph Neural Networks (GNN) and traditional deep learning models like HDGL.

\begin{figure}[h!]
    \centering
    \vspace*{-15pt}
    \includegraphics[width=0.47\textwidth]{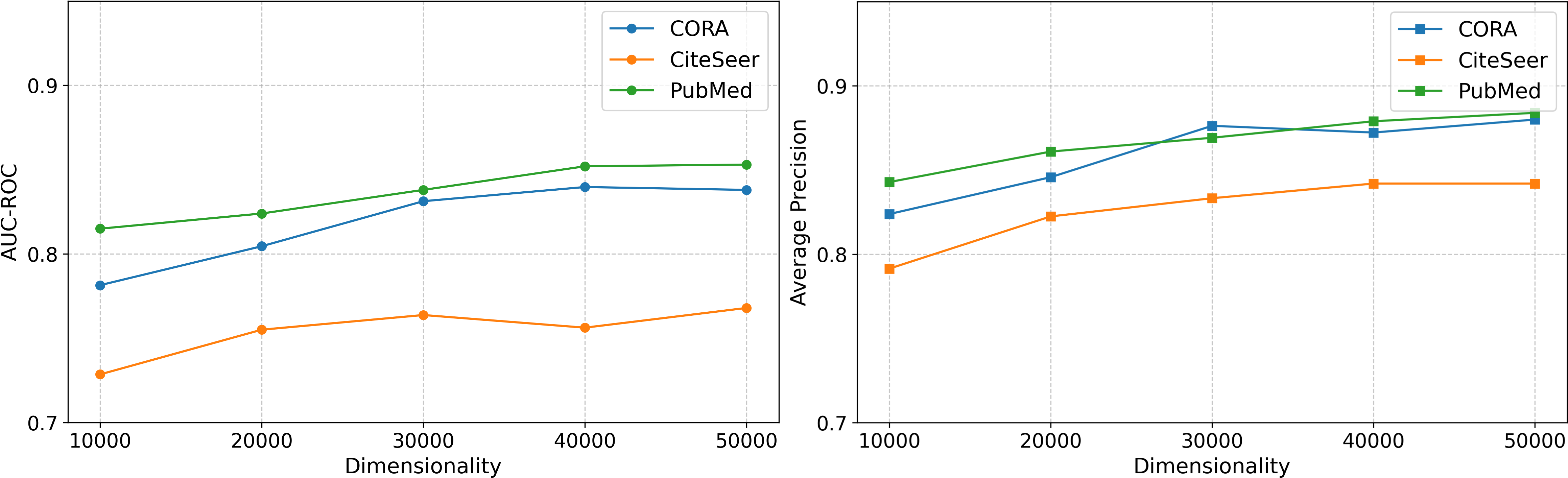}
    \vspace*{-0.4cm}
    \caption{Link Prediction Performance of HDGL under Various Dimensionality Configurations.}
    \vspace*{-2mm}
    \label{fig:link_performance_vs_dimensions}
    \Description{}
\end{figure}

\noindent\textbf{Link Prediction Performance. }
Even though HDGL representations are primarily tailored for transductive node classification, their performance for link prediction is comparable to that of DeepWalk and Spectral clustering; as seen from Table \ref{tab:link_pred_result}. However, they do not reach the same level of effectiveness as GAE and VGAE, which leverage GNN methods. Despite lacking iterative training like deep learning methods, HDGL still manages to achieve respectable results in link prediction. Moreover, unlike deep learning approaches that typically require distinct models for link and node classification, HDGL tackles both tasks within a unified framework. We also perform link prediction experiments by varying the dimensionality of the HDGL model. This exploration arises from observing a slight yet discernible variation in performance for link prediction tasks, unlike the more consistent outcomes observed in node classification experiments. As mentioned earlier, typically, dimensions ranging from 20,000 to 50,000 are sufficient for HD properties to hold. However, higher dimensionality generally leads to better performance. Nevertheless, beyond a certain threshold, the performance tends to plateau, as illustrated in Figure \ref{fig:link_performance_vs_dimensions}.

\begin{table}[h!]
    \centering
    \vspace*{-0.2cm}
    \caption{Link prediction results averaged over 10 Randomized Instantiations of HDGL and datasplits. Results for other models are from \citet{kipf2016variational_VGAE} paper.}
    \vspace*{-0.3cm}
    \label{tab:link_pred_result}
        \begin{tabular}{lcccccc}
            \toprule
            \multirow{2}{*}{\bf{Methods}}
                    & \multicolumn{2}{c}{\bf{Cora}} & \multicolumn{2}{c}{\bf{Citeseer}} & \multicolumn{2}{c}{\bf{PubMed}} \\
                & \bf{AUC} & \bf{AP} & \bf{AUC} & \bf{AP} & \bf{AUC} & \bf{AP} \\
            \midrule
            DeepWalk & 0.831  & 0.850  & 0.805  & 0.836  & 0.844  & 0.841 \\
            SC & 0.846 & 0.885 & 0.791 & 0.826 & 0.849 & 0.888 \\
            \midrule
            GAE & 0.910 & 0.920 & 0.895 & 0.899 & 0.964 & 0.965 \\
            VGAE & 0.914 & 0.926 & 0.908 & 0.920 & 0.944 & 0.947 \\
            \midrule
            HDGL & 0.849 & 0.880 & 0.768 & 0.842 & 0.853 & 0.884 \\
            \bottomrule
        \end{tabular}
\end{table}

\noindent\textbf{HDGL in the Class-Incremental Learning Setting}
Consider a scenario with a dynamic graph where nodes and edges remain static, but the new node labels appear over time i.e initially, only a subset of nodes is labeled, belonging to two classes. Over time, new nodes receive labels; thus, introducing new classes. This is a node labeling instance of the class-incremental learning problem \cite{van2022three,belouadah2021comprehensive,tian2024survey}. In this setting, deep learning methods typically require retraining on the data for previously learned classes along with the data for the new classes to avoid catastrophic forgetting of the previously learned classes \cite{tian2024survey}. This retraining incurs significant additional runtime and computation.

In contrast to deep neural networks, which necessitate retraining, Hyperdimensional Graph Learner (HDGL) operates without an iterative training phase. Instead, it constructs label hypervectors using predefined latent node representations. This approach eliminates the need for retraining, resulting in substantial savings in both runtime and computation costs.

To simulate incremental node classification on a fully labeled graph, we use BlogCatalog and Coauthor Physics Dataset. We first split the nodes into train/val/test as in \citet{shchur2018pitfalls}. We mask the train/val/test such that at time $ t=1 $ only the labels $\{1,2\}$ are available; and at $ t=2 $ only the labels $\{1,2,3\}$ are available; and so on such that at the final time step, all labels $\{1, \cdots, L\}$ are available.

To simplify matters for GNN models, we assume that the models are provided the list of $L$ possible labels although only a subset have been encountered in the training data. This allows the last layer of the GNN to have $L$ outputs. Also, we forego hyperpameter search for the GNN models to minimize the computational overhead.

In comparing HDGL against GCN and SGC in our experimental setup, we opted not to include the GAT model due to its longer runtime, as depicted in Figure \ref{fig:Run_time_compare}. As seen from Table \ref{tab:incremental_results}, we see that HDGL incurs a significant computational overhead only at timestep 1, necessitating the computation of node representations followed by inference. However, for subsequent timesteps , HDGL leverages previously computed node representations, requiring only inference. In contrast, deep learning models like GCN and SGC mandate retraining for each timestep to accommodate new labels, resulting in comparatively slower runtimes.

\section{Conclusion}
\label{sec:conc}
We introduced HDGL, a novel transductive learning algorithm for graphs using hyperdimensional representations. HDGL offers a computationally efficient alternative to graph neural networks by leveraging the \emph{injectivity} property of node representations from Graph Neural Networks (GNNs). It uses HD operators such as bundling and binding to aggregate information from the local neighborhood of each node. The resulting latent node representations support both node classification and link prediction tasks. Furthermore, HDGL eliminates the need for iterative training, making it ideal for class-incremental learning and applications requiring high accuracy models at lower computational cost and learning time compared to traditional graph neural network methods. The advantages and efficacy of HDGL are validated by comprehensive experiments comparing it with state-of-the-art GNN methods.

\begin{acks}
This work was funded in part by grants from the National Science Foundation (2226025), the National Center for Advancing Translational Sciences, and the National Institutes of Health (UL1 TR002014)
\end{acks}

\bibliographystyle{ACM-Reference-Format}
\balance
\bibliography{ref_wsdm}





\end{document}